%% file: main.tex
\documentclass[letterpaper, 10 pt, conference]{ieeeconf}
\IEEEoverridecommandlockouts 
\usepackage{graphics} % for pdf, bitmapped graphics files
\usepackage{amsmath} % assumes amsmath package installed
\usepackage{amssymb}  % assumes amsmath package installed
\usepackage{cite}
\usepackage{amsfonts}
\usepackage{bm}
\usepackage{breqn}

\usepackage{diagbox}
\usepackage{multirow}
\usepackage{multicol}
\usepackage{graphicx}
 \usepackage[subtle,tracking=normal]{savetrees}

\usepackage{balance}
\usepackage{caption}
\usepackage{subcaption}
\usepackage{makecell}

\usepackage[dvipsnames]{xcolor}
\definecolor{ao(english)}{rgb}{0.0, 0.5, 0.0}
\definecolor{green(pigment)}{rgb}{0., 0.75, 0.2}
\usepackage{soul}
\newcommand{\green}[1]{\color{green(pigment)}{#1}\color{black}}
\newcommand{\red}[1]{\color{red}{#1}\color{black}} % 

\usepackage{hyperref}
\usepackage[percent]{overpic}
\usepackage{xcolor}
\newcommand{\insertYoutubeLink}{\url{https://youtu.be/wpsbSMzIwgM}}

\title{\LARGE \bf
Visual CPG-RL: Learning Central Pattern Generators  \\for Visually-Guided Quadruped Locomotion
}

\author{Guillaume Bellegarda, Milad Shafiee, Auke Ijspeert% 
\thanks{
This research is supported by the Swiss National Science Foundation
(SNSF) as part of project No.197237. The authors are with the BioRobotics Laboratory, Ecole Polytechnique Federale de Lausanne (EPFL).
 {\tt \{firstname.lastname\}@epfl.ch}}
}

\begin{document}
\bstctlcite{MyBSTcontrol}
\maketitle
\thispagestyle{empty}
\pagestyle{empty}

\input{main_content}


%%%%%%%%%%%%%%%%%%%%%%%%%%%%%%%%%%%%%%%%%%%%%%%%%%%%%%%%%%%%%%%%%%%%%%%%%%%%%%%%
\bibliographystyle{IEEEtran}
\bibliography{refs}

\end{document}

%% file: main_content.tex
%%%%%%%%%%%%%%%%%%%%%%%%%%%%%%%%%%%%%%%%%%%%%%%%%%%%%%%%%%%%%%%%%%%%%%%%%%%%%%%%
% Abstract
%%%%%%%%%%%%%%%%%%%%%%%%%%%%%%%%%%%%%%%%%%%%%%%%%%%%%%%%%%%%%%%%%%%%%%%%%%%%%%%%
\begin{abstract}
We present a framework for learning visually-guided quadruped locomotion by integrating exteroceptive sensing and central pattern generators (CPGs), i.e. systems of coupled oscillators, into the deep reinforcement learning (DRL) framework. Through both exteroceptive and proprioceptive sensing, the agent learns to coordinate rhythmic behavior among different oscillators to track velocity commands, while at the same time override these commands to avoid collisions with the environment. We investigate several open robotics and neuroscience questions: 1) What is the role of explicit interoscillator couplings between oscillators, and can such coupling improve sim-to-real transfer for navigation robustness? 2) What are the effects of using a memory-enabled vs. a memory-free policy network with respect to robustness, energy-efficiency, and tracking performance in sim-to-real navigation tasks? 3) How do animals manage to tolerate high sensorimotor delays, yet still produce smooth and robust gaits? To answer these questions, we train our perceptive locomotion policies in simulation and perform sim-to-real transfers to the Unitree Go1 quadruped, where we observe robust navigation in a variety of scenarios. Our results show that the CPG, explicit interoscillator couplings, and memory-enabled policy representations are all beneficial for energy efficiency, robustness to noise and sensory delays of 90 \texttt{ms}, and tracking performance for successful sim-to-real transfer for navigation tasks. Video results can be found at \insertYoutubeLink.
\end{abstract}

%%%%%%%%%%%%%%%%%%%%%%%%%%%%%%%%%%%%%%%%%%%%%%%%%%%%%%%%%%%%%%%%%%%%%%%%%%%%%%%%
% Introduction
%%%%%%%%%%%%%%%%%%%%%%%%%%%%%%%%%%%%%%%%%%%%%%%%%%%%%%%%%%%%%%%%%%%%%%%%%%%%%%%%
\section{Introduction}
\label{sec:introduction}

Animals perform complex navigation tasks over variable terrain in the search for prey or to escape predators. In order to plan and execute such agile behaviors in unknown environments, exteroceptive sensing is necessary for both planning and control purposes (i.e.~anticipatory behavior vs. reactive behavior from ``blind'' walking). For both animals and robots, adding exteroception is both an opportunity (anticipation, planning) and a challenge (higher dimensional measurements, noise) for a control architecture.  On the other hand, animals process such high-dimensional information very quickly, and have internal mechanisms that share control between the spinal cord and higher control centers (e.g.~motor cortex and cerebellum). This avoids the concept that all motor commands come from higher control centers (i.e.~the biological parallel of current optimal control methods and learning-based policies). Towards a better biological parallel, in this work we represent higher control centers with an artificial neural network, which sends modulation signals to the Central Pattern Generator (CPG) in the spinal cord. The CPG is represented as a system of oscillators, and its states are modulated through feedback from both exteroceptive (i.e.~visual features) and proprioceptive (i.e.~base velocities, contact feedback) sensing to produce robust navigation policies. This framework enables us to investigate several scientific questions, and in particular we study 1) the role of interoscillator couplings in the spinal cord, 2) the role of a memory-enabled vs. a memory-free neural network as a higher control center, and 3) the effects of sensory delays on motor performance. The answers to these questions have not yet been confirmed from biology for animal navigation, nor addressed in robotics for robot obstacle avoidance. Here, we train and deploy 12 different navigation policies in over 250 sim-to-real experiments. Our results suggest that CPGs are useful intermediary control layers compared to direct joint control, that  memory-enabled networks perform better than memory-free networks, and that when vision is included, interoscillator couplings are useful vs.~no couplings.

\begin{figure}[!t]
    \centering
    \includegraphics[width=\linewidth,trim={0cm 2cm 0cm 0.15cm},clip]{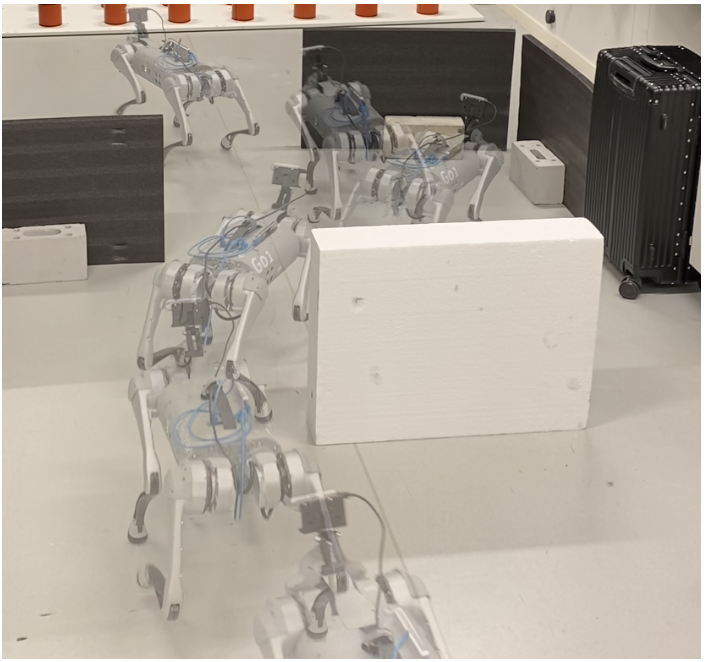}\\
    \vspace{-0.38em}
    \caption{Visual CPG-RL perceptive locomotion on Unitree Go1.}
    \label{fig:intro}
    \vspace{-2em}
\end{figure}

\vspace{-0.3em}
\subsection{Related Work} 
\subsubsection{Biology-Inspired Control} 
Central Pattern Generators are neural networks found in the spinal cord and brainstem of vertebrates which are capable of generating rhythmic patterns of muscle activity, such as those involved in walking, swimming, and flying~\cite{ijspeert2008}. For quadrupedal robots, CPGs have been used as a tool to both replicate and understand locomotion from several points of view, including the role of sensory input, reflexes, and mechanical design inspired from biology~\cite{Fukuoka2003,sprowitz2013cheetah}. While most works integrating sensory feedback with CPGs have focused on using proprioceptive sensing, incorporating cameras and other external sensory feedback have demonstrated walking over varying terrains~\cite{gay2013learncpg}, and responding to sudden obstacle appearances~\cite{saputra2021aquro,saputra2021combining}.

\subsubsection{Model-Based Control} 
Model Predictive Control (MPC) approaches have demonstrated that  using simplified dynamics models with filtered perception data as constraints during the optimization process can produce robust locomotion skills~\cite{kim2020vision,grandia2022perceptive,jenelten2022tamols,agrawal2022vision}. These works build an elevation map of the environment through one or more depth cameras~\cite{Fankhauser2018ProbabilisticTerrainMapping,Fankhauser2014RobotCentricElevationMapping}, and the resulting grid map~\cite{Fankhauser2016GridMapLibrary} is postprocessed for smoothing, inpainting, or plane segmentation, depending on the application. 

\subsubsection{Learning-Based Control} 
Learning-based control has also demonstrated impressive real-world capabilities, even for ``blind'' policies which have access only to proprioceptive sensing~\cite{tan2018minitaur,iscen2018policies,hwangbo2019anymal,kumar2021rma,lee2020anymal,chen2022learning,bellegardaIROS19TaskSpaceRL,bellegarda2020robust,bellegarda2022robust,shafiee2023manyquadrupeds}.  Similarly to recent MPC approaches, integrating perception into such learning-based control approaches can lead to even more robust locomotion. Demonstrative applications include combining occupancy maps with proprioceptive sensing for navigation~\cite{fu2022coupling}, obstacle avoidance with end-to-end training directly from pixels~\cite{yang2022learning,imai2021vision}, local navigation for challenging short time horizon tasks \cite{rudin2022advanced}, and crossing rough terrain dynamically by sampling height maps extracted from 3D Lidar~\cite{miki2022learning,rudin2022anymalisaac}. Other works have decoupled footstep planning processes from the control policy, either each trained with DRL~\cite{tsounis2020deepgait}, or leveraging MPC to track the footstep plan~\cite{gangapurwala2022rloc}. Learning a state representation of the environment from depth images and sending high-level commands to a separately trained DRL control policy also allows navigating cluttered environments~\cite{hoeller2021learning}. Additional works have used vision to demonstrate variable gap crossing capabilities\cite{yu2022visual,lee2022pi,margolis2022pixels,shafiee2023puppeteer,shafiee2023deeptransition}. 

In our previous work, we proposed CPG-RL~\cite{bellegarda2022cpgrl}, a framework for using deep reinforcement learning to directly learn the time-varying oscillator intrinsic amplitude and frequency for each oscillator which together forms a central pattern generator. We implemented the CPG network with one oscillator per limb, but without explicit couplings between oscillators as they did not prove beneficial for locomotion with only proprioceptive sensing.

\subsection{Contribution}
In this paper, we present a framework for incorporating terrain-awareness as exteroceptive feedback to the policy network which in turn modulates the CPG, enabling our quadruped to navigate efficiently in cluttered terrains. With our framework, we center our scientific investigation around three fundamental robotics and neuroscience questions: 
\begin{itemize}
    \item [1.] What is the role of explicit interoscillator couplings between oscillators, and can such coupling improve the sim-to-real transfer for perceptive locomotion tasks?
    \item [2.] What are the effects of using a memory-enabled vs. a memory-free policy network with respect to robustness, energy-efficiency, and tracking performance in the sim-to-real navigation task?
    \item [3.] Does the CPG (with/without coupling) increase robustness with respect to baselines when sensory delays are present? 
\end{itemize}

For question (1), we train perceptive locomotion policies with varying explicit coupling factors between oscillators in the dynamics equations. While such couplings between oscillators are known to exist in biological CPGs,  recent work has shown that they might not be as strong as previously thought~\cite{owaki2017quadruped,thandiackal2021emergence}, and that sensory feedback and descending modulation might play an important role in interoscillator synchronization. For the sim-to-real transfer, in contrast with our previous work on ``blind'' locomotion~\cite{bellegarda2022cpgrl}, we find that explicit coupling improves policy robustness when concatenating high-dimensional (and noisy)  exteroceptive inputs with the lower-dimensional proprioceptive sensing.

For question (2), we test different neural network architectures, specifically purely feedforward networks (MLPs) as well as memory-enabled networks (LSTMs), and find that the memory-enabled networks produce more robust and energy-efficient policies for sim-to-real transfers. 

For question (3), we compare and evaluate our method with a joint-space policy baseline and find that our architecture, and the CPG in particular, is able to maintain robustness to large sensory delays, comparable to those known to exist in similarly sized mammals~\cite{more2018scaling,ashtiani2021hybrid}.   

The rest of this paper is organized as follows. In Section~\ref{sec:method} we present Visual CPG-RL, including our design choices and integration of Central Pattern Generators and exteroceptive sensing into the deep reinforcement learning framework. In Section~\ref{sec:result} we discuss results and analysis from learning our controller and sim-to-real transfers for varying neural network architectures, specified interoscillator couplings, and sensory delays, and we give a brief conclusion in Section~\ref{sec:conclusion}.

%%%%%%%%%%%%%%%%%%%%%%%%%%%%%%%%%%%%%%%%%%%%%%%%%%%%%%%%%%%%%%%%%%%%%%%%%%%%%%%%%%%
% Method 
%%%%%%%%%%%%%%%%%%%%%%%%%%%%%%%%%%%%%%%%%%%%%%%%%%%%%%%%%%%%%%%%%%%%%%%%%%%%%%%%
\section{Learning Central Pattern Generators for Visually-Guided Locomotion}
\label{sec:method}

\begin{figure*}[!t]
      \centering
      \includegraphics[width=\linewidth]{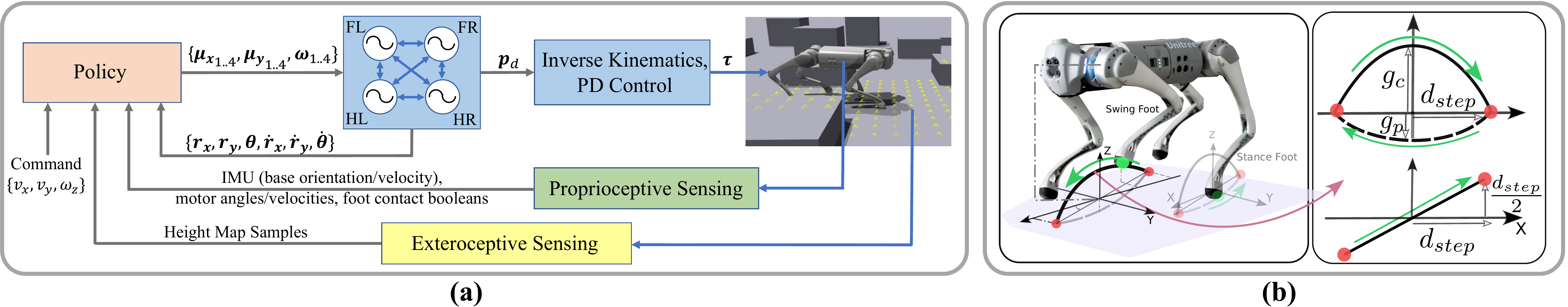}\\
      \vspace{-0.5em}
      \caption{\textbf{(a)}: Control architecture for learning central pattern generators for visually-guided quadruped locomotion. The observation consists of velocity commands, exteroceptive measurements, proprioceptive measurements, and the current CPG states, which the policy network uses to select CPG parameters $\mu_x$, $\mu_y$, and $\omega$ for each leg $i$ (Front Left (FL), Front Right (FR), Hind Left (HL), Hind Right (HR)). The resulting CPG states are mapped to desired foot positions $\bm{p}_d$, which are then converted to desired joint angles with inverse kinematics, and finally tracked with joint PD control to produce torques $\bm{\tau}$. The control policy selects actions at 100 Hz, and all other blocks operate at 1 kHz.
      \textbf{(b)}: Mapping CPG states to Cartesian foot positions. Left: feet path during swing and stance phases. Top right in the (vertical) XZ-plane: ground clearance ($g_c$), ground penetration ($g_p$), max step length ($d_{step}$) are design parameters, whereas CPG states $r_x$ and $\theta$ control amplitude and phase. Bottom right in the (horizontal) XY-plane: coordinating omnidirectional motion in the leg frame (arrow shows swing phase motion) with converged amplitude set points $\mu_x=2$, $\mu_y=1.25$, representing the full $d_{step}$ and $\frac{1}{2}d_{step}$, respectively.
      }
      \label{fig:control_diagram}
      \vspace{-1.5em}
\end{figure*}

In this section we describe our CPG-integrated deep reinforcement learning framework and design decisions for learning visually-guided locomotion controllers for quadruped robots. The agent receives exteroceptive and proprioceptive sensing measurements and the current CPG state as input, and learns to modulate and coordinate the CPG parameters for each limb to track velocity commands while avoiding collisions. A high-level control diagram is illustrated in Figure~\ref{fig:control_diagram}, and we explain all components below. 

%-------------------------------------CPG + Action Space-------------------------------------
\subsection{Central Pattern Generators and Action Space}
\label{sec:action_space}
Based on our previous work~\cite{bellegarda2022cpgrl}, in this work we propose the following amplitude-controlled phase oscillators to coordinate locomotion between each limb $i$: 
\begin{align}
\ddot{r}_{x_i} &= a_x \left(\frac{a_x}{4} \left(\mu_{x_i} - r_{x_i} \right) - \dot{r}_{x_i} \right) \label{eq:rl_rx}\\
\ddot{r}_{y_i} &= a_y \left(\frac{a_y}{4} \left(\mu_{y_i} - r_{y_i} \right) - \dot{r}_{y_i} \right) \label{eq:rl_ry}\\
\dot{\theta}_i &= \omega_i  + \frac{1}{2}\sum_{j}^{} (r_{x_j} + r_{y_j}) w_{ij} \sin(\theta_j - \theta_i - \phi_{ij}) \label{eq:rl_theta2}
\end{align}
where $r_{x_i}$ and $r_{y_i}$ are the current amplitudes of the oscillator, $\theta_i$ is the phase of the oscillator, $\mu_i$ and $\omega_i$ are the intrinsic amplitude and frequency, $a$ is a positive constant representing the convergence factor. Couplings between oscillators are defined by the weights $w_{ij}$ and phase biases $\phi_{ij}$.

Compared with previous work that typically uses a single amplitude and phase variable, here we have a third state variable to represent amplitude in the $y$ direction. The agent can then learn to coordinate both amplitudes for omnidirectional locomotion (see Figure~\ref{fig:control_diagram}-b), and we update the coupling to reflect both $x$ and $y$ components. In particular, this mapping is different from our previous work~\cite{bellegarda2022cpgrl}, where we had added an extra phase variable to orient the foot, instead of the new proposed amplitude. We find this new mapping is able to achieve higher returns during training.

Our action space provides an interface for the agent to directly modulate the intrinsic oscillator amplitudes and phases, by learning to modulate $\mu_{x_i}$, $\mu_{y_i}$, and $\omega_i$ for each leg. This allows the agent to adapt each of these states online in real-time depending on sensory inputs, in contrast with the more traditional CPG approach of optimizing for a specific fixed open-loop gait. Thus, for the omnidirectional perceptive locomotion task, our action space can be summarized as $\mathbf{a} = [\bm{\mu}_x, \bm{\mu}_y, \bm{\omega}] \in \mathbb{R}^{12}$. The agent selects these parameters at 100 Hz, and we use the following action space ranges during training: $\mu_x, \mu_y \in [1, 2]$, $\omega \in [0, 4.5]$ Hz, and $a_x=a_y=50$.  

The oscillator states are mapped to joint commands by first computing corresponding desired foot positions, and then calculating the desired motor angles with inverse kinematics. This is an approximation of the typical two layers found in mammalian CPGs, with one rhythm generating layer, and one pattern formation layer~\cite{mccrea2008organization}, with here the pattern formation layer implementing the inverse kinematics. The desired foot position coordinates are computed as:
\begin{align}
x_{i,\text{foot}} &= -d_{step} f(r_{x_i}) \cos(\theta_i) \\
y_{i,\text{foot}} &= \phantom{-}d_{step} f(r_{y_i}) \cos(\theta_i) \\
z_{i,\text{foot}} &= \begin{cases}
    -h + g_c\sin(\theta_i) & \text{if } \sin(\theta_i) > 0 \\
    -h + g_p\sin(\theta_i) & \text{otherwise}
\end{cases}
\end{align}
where $d_{step}$ is the maximum step length, $h$ is the robot height, $g_c$ is the max ground clearance during swing,  $g_p$ is the max ground penetration during stance, and  $f(r) = 2 \frac{(r - \mu_{\text{min}})}{(\mu_{\text{max}} - \mu_{\text{min}})} - 1$. In this mapping, since $r_{x_i}$ and $r_{y_i}$ each vary between $\mu_{\text{min}}=1$ and $\mu_{\text{max}}=2$, the foot can vary within $\pm d_{step}$ in both $x$ and $y$ directions in each leg frame.

Figure~\ref{fig:control_diagram}-b illustrates the foot trajectories for a set of these parameters, which greatly simplifies specifying behaviors that are challenging to learn when directly learning joint commands. As in~\cite{bellegarda2022cpgrl}, we sample $h$, $g_c$, and $g_p$ during training so the agent can learn to locomote with varying base heights, swing foot ground clearances, and stance foot ground penetrations. The agent does not receive any explicit observation of these parameters, and the user can specify each of these parameters during deployment.

%-------------------------------------Observation Space-------------------------------------
\subsection{Observation Space}
\label{sec:obs_space}
Our observation space consists of velocity commands and measurements available with proprioceptive sensing as in~\cite{bellegarda2022cpgrl}, as well as exteroceptive sensing. The exteroceptive measurements consist of querying a terrain height map at a $17 \times 11$ grid spaced at intervals of 0.1 \texttt{m} around the robot base (as shown by the yellow dots in Figure~\ref{fig:control_diagram}-a). In simulation, the ground truth terrain height data is known, and on hardware such a grid can be estimated, for example by using depth cameras to build an elevation map~\cite{Fankhauser2018ProbabilisticTerrainMapping,Fankhauser2014RobotCentricElevationMapping}, and then querying the resulting postprocessed grid map~\cite{Fankhauser2016GridMapLibrary}. 

The proprioceptive sensing includes the body state (orientation, linear and angular velocities), joint state (positions, velocities), and foot contact booleans. The last action chosen by the policy network and CPG states $\{\bm{r}_x,\bm{\dot{r}}_x,\bm{r}_y,\bm{\dot{r}}_y,\bm{\theta},\bm{\dot{\theta}}\}$ are concatenated to the exteroceptive and proprioceptive measurements. While the exteroceptive and proprioceptive sensing are subject to measurement noise from onboard sensors, the CPG states are always known both in simulation and during hardware experiments, easing the sim-to-real transfer.  Sensorimotor delays are present in both animals and robots, and the exteroceptive (vision) and proprioceptive sensing loops operate at different frequencies. We investigate the effects of sensory delays in Section~\ref{sec:sensory_delay}.

%------------------------------------- Reward-------------------------------------
\subsection{Reward Function} 
\label{sec:reward_function}

Our reward function primarily rewards tracking body linear and angular velocity commands in the base frame ($x$ and $ y$ directions, as well as yaw rate $\omega_{b,z}^{*}$). We also add terms to minimize other undesired base velocities (vertical oscillations in the base $z$ direction, and base roll and pitch rates). To minimize energy consumption, we penalize the total power. The terms and respective weights are summarized in Table~\ref{tab:reward}. Similarly to~\cite{bellegarda2022cpgrl}, we do not need to add any reward terms beyond those fully specifying the base motion behavior. Due to the variable height terrain in the training environment, however, it is not possible to track all velocity commands at all times (i.e. if there are obstacles in the way). As shown in the Table, we more heavily weight the forward velocity reward term $v_{b,x}^{*}$ so the policy learns to prefer deviations to the other velocity commands/penalties (i.e. to turn if approaching an obstacle head-on). 

\begin{table}[tpb]
\centering
\caption{Reward function terms. $(\cdot)^{*}$ represents a desired command, and $f(x) := \exp{(-\frac{||x||^2}{0.25}) } $. $dt=0.01$ is the control policy time step. 
}
\vspace{-0.3em}
\begin{tabular}{ c c c }
Name & Formula & Weight \\
\hline
Linear velocity tracking $v_{b,x}^{*}$ & $f(v_{b,x}^{*} - v_{b,x})$ & $3 dt$ \\
Linear velocity tracking $v_{b,y}^{*}$ & $f(v_{b,y}^{*} - v_{b,y})$ & $0.75 dt$ \\
Angular velocity tracking $\omega_{b,z}^{*}$ & $f(\omega_{b,z}^{*} - \omega_{b,z})$ & $0.5dt$ \\
\hline
Linear velocity penalty $v_{b,z}$ & $-v_{b,z}^2$ & $2dt$ \\
Angular velocity penalty $\bm{\omega}_{b,xy}$ & $-||\bm{\omega}_{b,xy}||^2$ & $0.05dt$ \\
Power & $-|\bm{\tau} \cdot \dot{\bm{q}}| $  & $0.001dt$ \\
\hline
\end{tabular} \\
\label{tab:reward}
\vspace{-2.2em}
\end{table}

%-------------------------------------NN-------------------------------------
\subsection{Neural Network Architectures}
\label{sec:nn}

To address our question about the importace of memory (question 2), we consider two different neural network architectures to map the concatenated proprioceptive and exteroceptive sensing observation to actions modulating the intrinsic oscillator amplitudes and phases. The first is a purely feedforward network, or multi-layer perceptron (MLP), consisting of three hidden layers of [512, 256, 128] hidden units per layer. The second architecture is a memory-enabled network, consisting of a Long Short-Term Memory (LSTM) layer of 512 hidden units, followed by two fully connected layers of [256, 128] hidden units. The memory-enabled network has a better biological parallel, and we also anticipate this will provide better robustness for the sim-to-real transfer in the event of noisy measurements and latency.

%-------------------------------------Training Details-------------------------------------
\subsection{Training Details}
\label{sec:training_details}

We use Isaac Gym and PhysX as our training environment and physics engine~\cite{isaacgym,rudin2022anymalisaac}, and the Unitree Go1 quadruped~\cite{unitreeGO1}. This framework has high throughput, enabling us to simulate 4096 Go1s in parallel on a single NVIDIA RTX 3090 GPU. We use the Proximal Policy Optimization (PPO) algorithm~\cite{ppo} to train the policy, with the same hyperparameters as in~\cite{bellegarda2022cpgrl}. Similarly to~\cite{rudin2022anymalisaac,bellegarda2022cpgrl}, this framework allows us to learn control policies within minutes. 

During training, we reset the environment for an agent if the base or a thigh comes in contact with the terrain (i.e.~with either a box or the ground, so the agent learns to avoid collisions), or if the episode length reaches 20 seconds. We employ a terrain curriculum starting from flat terrain, to random boxes of varying widths (0.4 to 2 \texttt{m}) and heights (0.1 to 1 \texttt{m}) so the agent can learn to avoid obstacles in a variety of scenarios. With each reset, we sample new parameters $h$ and $g_c$ for mapping the oscillator states to motor commands, allowing the agent to learn continuous locomotion behavior at varying body heights and step heights. New velocity commands $\{v_{b,x}^{*},v_{b,y}^{*},\omega_{b,z}^{*}\}$ are sampled every 5 seconds, though as explained in Section~\ref{sec:reward_function}, these cannot be perfectly tracked due to the presence of obstacles, which forces the agent to learn to deviate from commands to avoid terminating the episode. We also apply domain randomization on the physical mass properties and coefficient of friction, as summarized in Table~\ref{table:dyn_rand}. An external push of up to 0.5 \texttt{m/s} is applied in a random direction to the base every 15 seconds. While no noise is added to the proprioceptive measurements, we add Gaussian noise to the exteroceptive measurements with a standard deviation of 0.1.

The policy network outputs modulation signals at 100 Hz, and the torques computed from the mapped desired joint positions are updated at 1 kHz. The equations for each of the oscillators (Equations~\ref{eq:rl_rx}-\ref{eq:rl_theta2}) are thus also integrated at 1 kHz. During training we re-sample joint PD controller gains at each environment reset as described in Table~\ref{table:dyn_rand}, but during deployment we use $K_p=100,\ K_d=2$. 

\begin{table}[tpb]
\centering
\caption{Randomized parameters during training and their ranges.}
\vspace{-0.4em}
\begin{tabular}{ c c c c }
Parameter & Lower Bound & Upper Bound & Units \\
\hline 
$v_{b,x}^{*}$    & -0.6 & 0.6 & \texttt{m/s}\\
$v_{b,y}^{*}$    & -0.4 & 0.4 & \texttt{m/s}\\
$\omega_{b,z}^{*}$ & -0.8 & 0.8 & \texttt{rad/s}\\
\hline
Joint Gain $K_p$ & 55 & 100 & - \\
Joint Gain $K_d$ & 0.7 & 2.5 & - \\
\hline
Mass (each body link) & 70 & 130 & \%\\
Added base mass & 0  & 5 & $kg$ \\
Coefficient of friction & 0.3 & 1 & -\\
\hline
\end{tabular} \\
\label{table:dyn_rand}
\vspace{-2.4em}
\end{table}

%-------------------------------------sim2real-------------------------------------
\subsection{Sim-to-Real Transfer}
As in~\cite{kim2020vision}, we mount two Intel RealSense cameras (D435i and T265) on the Unitree Go1 quadruped~\cite{unitreeGO1}. The D435i is mounted forward-facing to provide point clouds to elevation mapping software~\cite{Fankhauser2018ProbabilisticTerrainMapping,Fankhauser2014RobotCentricElevationMapping} to construct a map of the area surrounding the robot. The T265 provides high accuracy localization. To run the policy, we query this map at the same $17 \times 11$ points around the robot as seen during training (i.e.~yellow dots in Figure~\ref{fig:control_diagram}), and concatenate the proprioceptive sensing measurements as read by the Unitree sensors, along with the CPG states and previous actions. 

%%%%%%%%%%%%%%%%%%%%%%%%%%%%%%%%%%%%%%%%%%%%%%%%%%%%%%%%%%%%%%%%%%%%%%%%%%%%
% Results
%%%%%%%%%%%%%%%%%%%%%%%%%%%%%%%%%%%%%%%%%%%%%%%%%%%%%%%%%%%%%%%%%%%%%%%%%%%%
\section{Experimental Results and Discussion}
\label{sec:result}

In this section we report and discuss results from learning visually-guided locomotion controllers with Visual CPG-RL. Sample snapshots of one of our perceptive locomotion policy deployments are shown in Figure~\ref{fig:intro}, and the reader is encouraged to watch the supplementary video for clear visualizations of the discussed experiments. Our experiments are designed to investigate the three questions detailed in the introduction. 

\begin{figure}
    \centering
    \includegraphics[width=0.97\linewidth,trim={4.2cm 7cm 4cm 7cm},clip]{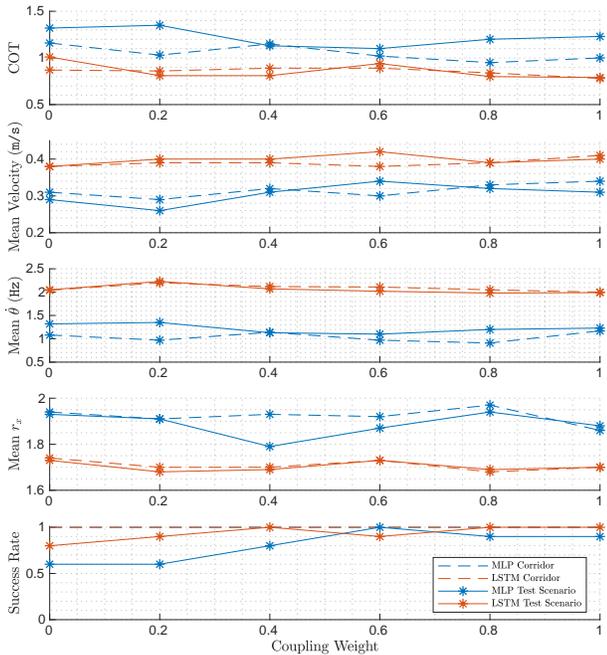}\\
    \vspace{-0.5em}
    \caption{Sim-to-real tracking performance of $v_{b,x}^{*} = 0.35$ \texttt{m/s} in a wide corridor (dashed line) and on a test navigation environment involving both left and right turns (solid lines), with both MLP and LSTM policies trained with varying coupling weights ($w_{i,j}$ in Equation~\ref{eq:rl_theta2}). Each data point represents the mean of 10 trials. From top to bottom, we present the mean Cost of Transport (COT), the quadruped base mean velocity, the mean frequency across all oscillators ($\dot{\theta}$), the mean amplitude $r_x$ correlated with the mean step length, and the success rate denoting avoidance of obstacle collisions or falls.}  
    \label{fig:sim2real_comp}
    \vspace{-2.5em}
\end{figure}

%------------------------------------------------------------------- 
\subsection{Role of Policy Architecture and Interoscillator Coupling}

We train policies with both neural network architectures (MLP and LSTM) described in Section~\ref{sec:nn}. For each architecture, we train separate policies with increasingly strong interoscillator coupling weights, namely $w_{i,j} = \{0,0.2,0.4,0.6,0.8,1.0\}$ in Equation~\ref{eq:rl_theta2}, for a total of 6 MLP policies and 6 LSTM policies. We train several policies with different random seeds within each of these categories, and select the one achieving the highest return for deployment. We use a trot gait coupling matrix, since without coupling (i.e.~$w_{i,j}=0$), all policies learn an approximate trot gait. 

%-------------------------------------------------------------------  corridor
\subsubsection{Corridor Test} We first test all policies in a 1.7 \texttt{m} wide corridor and command a forward velocity of $v_{b,x}^{*} = 0.35$ \texttt{m/s} for 10 seconds. This test is done to ensure successful sim-to-real transfers on (mostly) flat terrain, and mean results from 10 rollouts of each policy are shown by the dashed lines in Figure~\ref{fig:sim2real_comp}. All policies successfully locomote with 100\% success rate for both MLP and LSTM architectures, and with all varying coupling strengths. However, the LSTM policies have a lower Cost of Transport for all couplings. When comparing the CPG state mean amplitudes and frequencies, the LSTM policies locomote with a higher leg frequency but lower step length than the MLP policies. Due to having memory, the LSTM policies are able to better optimize quantities like stride length and stride frequency, leading to better energy-efficiency. The LSTM policies also track the commanded velocities slightly more closely, and interestingly overshoot the command, compared with the MLP policies which are consistently slower. 

\begin{figure}[!t]
    \centering
    \includegraphics[width=0.97\linewidth]{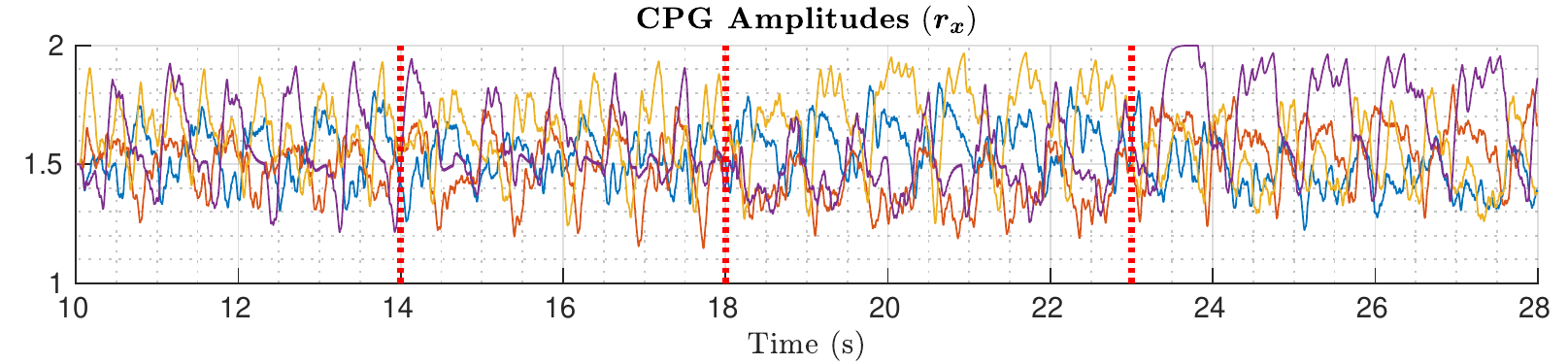} \\
    \vspace{0.2em}
    \includegraphics[width=0.97\linewidth]{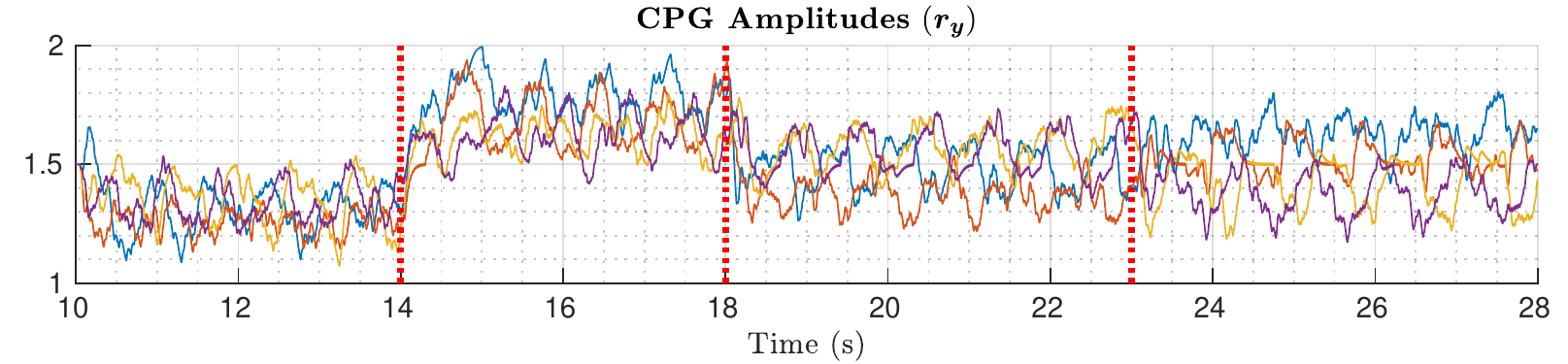} \\
    \vspace{0.2em}
    \includegraphics[width=0.95\linewidth]{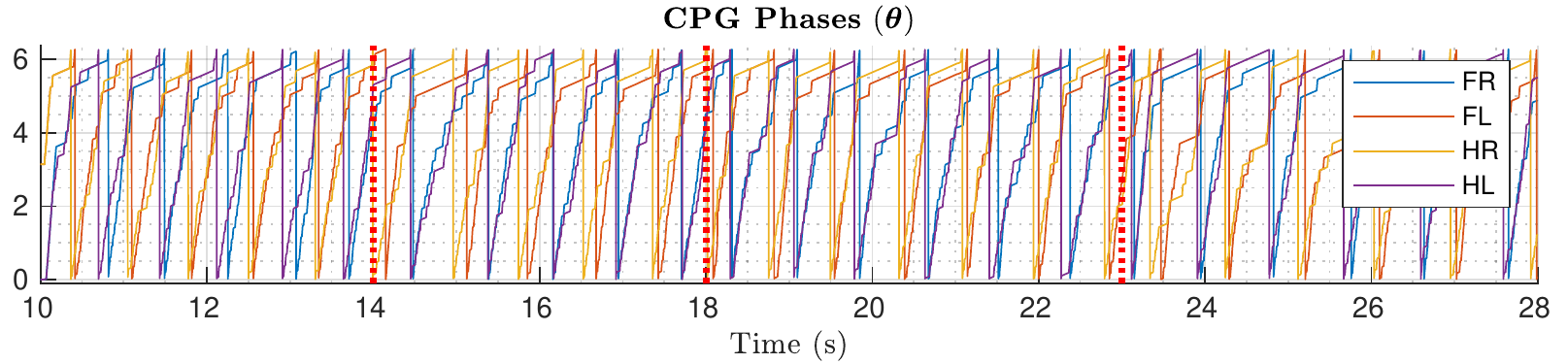} \\
    \vspace{-0.5em}
      \caption{CPG states during omnidirectional commands: $v_{b,y}^{*} = 0.4$ \texttt{m/s} from 10-14 \texttt{s}, $v_{b,y}^{*} = -0.4$ \texttt{m/s} from 14-18 \texttt{s}, $\omega_{b,z}^{*} = 0.7$ \texttt{rad/s} from 18-23 \texttt{s}, and $\omega_{b,z}^{*} = -0.7$ \texttt{rad/s} from 23-28 \texttt{s}.  The $y$ amplitudes $r_y$ produce locomotion for the system for the lateral commands. For turning in place, we observe coordination between both $x$ and $y$ amplitudes.   }
      \label{fig:y_wz}
      \vspace{-1.5em}
\end{figure}

%------------------------------------------------------------------- navigation
\subsubsection{Navigation Test}
\label{sec:nav_test}
We next test all policies in a navigation environment which requires both left and right turns in order to avoid obstacles. We define a failure as the robot colliding with an obstacle or falling down. We again command a forward velocity of $v_{b,x}^{*} = 0.35$ \texttt{m/s} for 10 seconds, and mean results from 10 rollouts of each policy are summarized by the solid lines in Figure~\ref{fig:sim2real_comp}. In order to avoid the obstacles, the agent is forced to violate at least one of the 0 velocity commands for $v_{b,y}^{*}$ and $\omega_{b,z}^{*}$. In these tests, for both the MLP and LSTM policies, we observe that coupling is beneficial for successful navigation of the terrain. The explicit coupling helps to reject noise and latency associated with the high-dimensional (relative to the proprioceptive) exteroceptive measurements. The LSTM policies also prove to be overall more robust than the MLP policies. 

%------------------------------------------------------------------- CPG state modulation
\subsection{CPG State Modulation for Omnidirectional Locomotion}
We investigate how the agent modulates the CPG states to produce omnidirectional locomotion on flat terrain. Figure~\ref{fig:y_wz} shows the CPG states when commanded to move laterally in the body $y$  directions, $v_{b,y}^{*} =\pm 0.4$ \texttt{m/s}, and then turn in place in both directions, $\omega_{b,z}^{*} = \pm 0.7$ \texttt{rad/s}. The video shows that the gait is smooth and consistent for the entirety of the motion, at a much lower frequency than the default Go1 controller. The lateral motions show the amplitudes $y$ selecting the expected directions, and coordination can be seen between both $x$ and $y$ amplitudes for turning in place. An approximate trot gait can be observed throughout all commands. For turning left, for example, the hind right foot has the largest amplitude in the $x$ direction, which combined with the $y$ amplitude moving mostly right in the body frame, turns the system in the expected direction. The reverse is true for turning right, where the hind left foot has the largest amplitude in the $x$ direction, and combines with the $y$ amplitude now moving mostly left in the body frame. 

In the video, we also show how the CPG states are modulated during a navigation test involving both left and right turns through exteroceptive feedback as discussed in Section~\ref{sec:nav_test}.

\begin{table*}[t!]
\centering
\caption{
Success rate, mean body angular velocity $\bar{\omega}$, and mean joint accelerations $\bar{\bm{\ddot{q}}}$ for 100 quadrupeds tracking  $v_{b,x}^{*} = 0.35$ \texttt{m/s} for 40 seconds with increasing sensory delays on the test environment of Figure~\ref{fig:sim_comp_ig}. High angular velocities and joint accelerations correspond to shaky and non-optimal locomotion.  
}
\vspace{-0.5em}
\begin{tabular}{ | c | c | c || c | c | c || c | c  | c  || c | c | c || c | c | c | }
\hline
& & & \multicolumn{12}{c|}{Sensory Delay (s) }  \\
 \hline
& & & \multicolumn{3}{c||}{0} & \multicolumn{3}{c||}{0.03} & \multicolumn{3}{c||}{0.06} & \multicolumn{3}{c |}{0.09} \\
 \hline
Method & NN & $w_{ij}$ & \thead{Success\\ Rate} & $\bar{\omega}$ & $\bar{\bm{\ddot{q}}}$ & \thead{Success\\ Rate} & $\bar{\omega}$ & $\bar{\bm{\ddot{q}}}$ & \thead{Success\\ Rate} & $\bar{\omega}$ & $\bar{\bm{\ddot{q}}}$ & \thead{Success\\ Rate} & $\bar{\omega}$ & $\bar{\bm{\ddot{q}}}$ \\
\hline
\hline
\multirowcell{4}{ Visual \\ CPG-RL }   & \multirowcell{2}{MLP} & 0 & \green{\bf{1}} & 0.26 & 0.85  & \green{\bf{1}}  &  0.26 & 0.84 & \green{\bf{1}} & \green{\bf{0.36}} & \green{\bf{0.86}} & 0.55 & \green{\bf{0.56}} & \green{\bf{0.92}} \\
\cline{3-15}
 &  & 1 & \green{\bf{1}} & \green{\bf{0.25}} & \green{\bf{0.83}} & \green{\bf{1}} &  \green{\bf{0.25}} & \green{\bf{0.83}} & \green{\bf{1}} & 0.38 & \green{\bf{0.86}} & \green{\bf{0.71}}  & \green{\bf{0.57}} & \green{\bf{0.97}} \\
\cline{2-15}
 & \multirowcell{2}{LSTM} & 0 & \green{\bf{1}} & 0.28 & 0.89 & \green{\bf{1}} & 0.28 & 0.90 & \green{\bf{1}} & 0.46 & 1.04 & \red{0} & 0.55 & 1.25 \\
\cline{3-15}
 &  & 1 & \green{\bf{1}} & 0.28 & 0.85 & \green{\bf{1}} & 0.28 & 0.85 & \green{\bf{1}} & 0.47 & 1.02 & \red{0} & 0.77 & 1.23   \\
\hline 
\hline
\multirowcell{2}{ Joint PD } & MLP & - & 0.96 & \green{\bf{0.25}} & 0.90 & 0.97 & \green{\bf{0.25}} & 0.90 & 0.98 & 0.43 & \red{3.98} & \red{0.08} & \red{0.88} & \red{2.26} \\
\cline{2-15}
 & LSTM & - & 0.97 & \green{\bf{0.25}} & 1.02 & 0.98 & \green{\bf{0.25}} & 1.01 & 0.98 & 0.43 & \red{1.25} & \red{0} & \red{0.92} & \red{2.20} \\
\hline
\end{tabular} 
\label{table:ablation1}
\vspace{-2em}
\end{table*}

%------------------------------------------------------------------- sensory delay
\subsection{Sensory Delay Study}
\label{sec:sensory_delay}

While in simulation it is possible to query any sensory information at any rate, real-world sensing will inevitably have latency issues, especially for vision-based systems. Although the control policy network is queried at 100 Hz, the terrain map on the hardware is built and updated at a lower frequency, meaning inevitable sensory delays and latency compared with in simulation. Several biological studies have shown that animals locomote with sensorimotor delays on the order of tens of milliseconds, increasing for larger mammals~\cite{more2018scaling}. Using the same scaling equation for the Go1 quadruped, which has a mass of 12 \texttt{kg}, would give an expected sensorimotor delay of $31 * 12^{0.21} = 52.24$ \texttt{ms} between stimulus onset and peak muscle force. 

\begin{figure}[t]
    \centering
    \vspace{0.2em}
    \includegraphics[width=\linewidth]{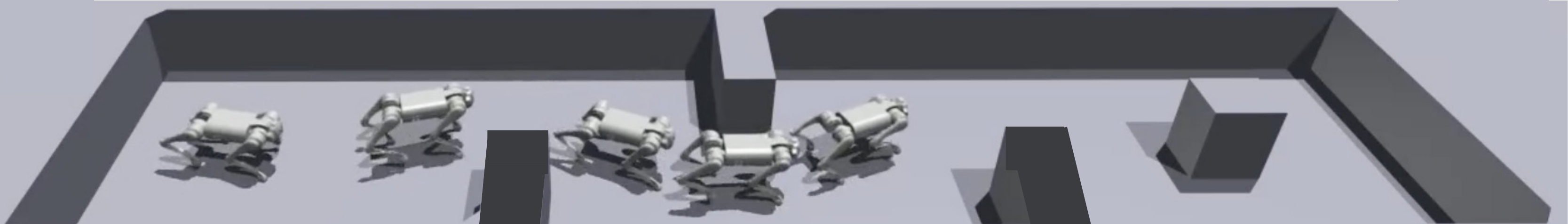}\\
    \vspace{-0.35em}
    \caption{Simulation test environment involving both left and right turns, as well as turning around an obstacle, as in the hardware experiments.  }
    \label{fig:sim_comp_ig}
    \vspace{-2.4em}
\end{figure}

In this section, we compare the effects of introducing such sensory delays on both the proprioceptive and exteroceptive measurements with our hierarchical biology-inspired architecture. As a baseline and to study the usefulness of integrating the CPG that exists in vertebrates with different neural networks and coupling weights, we train joint PD policies with a similar reward function to~\cite{rudin2022anymalisaac}. We note that the reward function for the joint PD baseline is much more complex and has to be significantly tuned, in contrast with Visual CPG-RL, in order to get similarly natural-looking navigation policies. The policies are all trained in the same environment described in Section~\ref{sec:training_details}. 

We evaluate Visual CPG-RL and the joint PD baseline on a test environment consisting of left and right turns, as well as an obstacle to move around, where the agent is free to choose to go left or right before coming back the way it came, as shown in Figure~\ref{fig:sim_comp_ig}. We let 100 agents for each method navigate in this terrain, and define a success as tracking the desired velocity of $v_{b,x}^{*} = 0.35$ \texttt{m/s} for 40 seconds without falling or colliding with the environment. 

Table~\ref{table:ablation1} summarizes results of testing the trained Visual CPG-RL policies with both neural network architectures (MLP and LSTM) and coupling weights $w_{i,j} = \{0,1\}$ (in Equation~\ref{eq:rl_theta2}), as well as the joint PD baseline with both NN architectures. We can observe that all policies perform well without any sensory delays. Notably, the Visual CPG-RL architecture performs almost identically through delays of 60 \texttt{ms}, which is greater than the expected animal-estimated sensorimotor delay of 52.24 \texttt{ms}. As the observations are increasingly delayed to 90 \texttt{ms}, the joint PD policies cannot produce any successful locomotion, and behave erratically with high body angular velocities and large joint angular accelerations (red values in the Table). In contrast, we note that Visual CPG-RL MLP policies perform well, and much better than the LSTM policies, whose internal states are mismatched with the new observation latency compared with the training environment.  We also observe that coupling appears to help the MLP policies perform better with increasing sensory delays. We note that the Visual CPG-RL policies are consistently smoother, as evidenced by the mean body angular velocity and joint accelerations being lower than the joint PD baseline, and are qualitatively more natural with increasing delays. 

%%%%%%%%%%%%%%%%%%%%%%%%%%%%%%%%%%%%%%%%%%%%%%%%%%%%%%%%%%%%%%%%%%%%%%%%%%%%%%%%
% Conclusion
%%%%%%%%%%%%%%%%%%%%%%%%%%%%%%%%%%%%%%%%%%%%%%%%%%%%%%%%%%%%%%%%%%%%%%%%%%%%%%%%
\section{Conclusion}
\label{sec:conclusion}

In this work we have presented Visual CPG-RL, a framework for learning perceptive quadruped locomotion by integrating central pattern generators and exteroceptive sensing into the deep reinforcement learning framework. The agent learns to modulate the intrinsic oscillator amplitudes and frequencies to coordinate rhythmic behavior among limbs to track omnidirectional velocity commands, while also learning to deviate from the commands in order to avoid collisions with obstacles. We represented higher control centers in the brain with an artifical neural network (ANN), and showed that memory-enabled networks (i.e.~LSTMs) provide higher robustness and encode more energy-efficient policies than feedforward networks (MLPs) for real-world navigation tasks which may have high-dimensional inputs, noise, and latency concerns (question 2). 
The ANN sends modulation signals to a system of oscillators in the spinal cord (CPG), and explicit couplings within the CPG dynamics equations proved to be beneficial in terms of robustness and stability for the sim-to-real navigation task (question 1). Compared with our previous work which did not investigate  interoscillator couplings for ``blind'' locomotion~\cite{bellegarda2022cpgrl}, these new results suggest the reason direct coupling is present in animals, which rely heavily on vision, and gives robots the ability to learn and deploy robust, adaptive, and efficient policies when subject to noisy high-dimensional inputs. From a robotics perspective, coupling between oscillators improves stability, at the cost of potential less flexibility (i.e.~it may be more difficult to recover from a push or other disturbance with strong and/or speed-dependent coupling~\cite{berendes2016speed}). Moreover, we studied the effects of sensory delays on policy robustness of each of the proposed neural network architectures and coupling weights, and found that the CPG is beneficial for suppressing sensory latency, and the performance was inline with estimated sensorimotor delays for a comparably sized animal~\cite{more2018scaling} (question 3). Future work will focus on adding additional states to the CPG to account for rough terrain, as well as compare with joint-space CPGs, towards a better understanding of the biological parallels in learning legged locomotion.